\documentclass[10pt, twoside]{article}
\usepackage{geometry}
\usepackage[utf8]{inputenc}
\usepackage{amsmath}
\usepackage{graphicx}
\usepackage{hyperref}
\usepackage{booktabs} 
\usepackage{subcaption}
\usepackage{float}
\usepackage{datetime}
\usepackage{xcolor}
\usepackage[figurename=Figure]{caption}
\usepackage{setspace}

\begin{document}

  %  \item \textbf{SI Conversion Factors (contingent):} Include a conversion factor table if only one system of units (customary or metric) is used throughout the paper. Include conversion factors for all units used, including those usedin figures and tables.  
  
\begin{center}
    {\huge Risk Analysis of Flowlines in the Oil and Gas Sector: A GIS and Machine Learning Approach}
    \vspace{5mm}
    
    \textbf{I.~Chittumuri$^{1}$,} \textbf{N.~Alshehab$^{2}$,} \textbf{R.~J.~Voss$^{1}$,} \textbf{L.~L.~Douglass$^{1}$,} \textbf{S.~Kamrava$^{2}$,} \textbf{Y.~Fan$^{2}$,} \textbf{J.~Miskimins$^{2}$,} \textbf{W.~Fleckenstein$^{3}$,} and \textbf{S.~Bandyopadhyay$^{1*}$} \vskip1em    $^{1}$Department of Applied Mathematics and Statistics, Colorado School of Mines\\
    $^{2}$Department of Petroleum Engineering, Colorado School of Mines\\
    $^{3}$Office of Global Initiatives and Business Development, Colorado School of Mines
\end{center}

*Corresponding author; email: \url{sbandyopadhyay@mines.edu}
\vspace{3mm}

\vspace{5mm}

\section*{Summary}
This paper presents a risk analysis of flowlines in the oil and gas sector using Geographic Information Systems (GIS) and machine learning (ML). Flowlines, vital conduits transporting oil, gas, and water from wellheads to surface facilities, often face under-assessment compared to transmission pipelines. This study addresses this gap using advanced tools to predict and mitigate failures, improving environmental safety and reducing human exposure. Extensive datasets from the Colorado Energy and Carbon Management Commission (ECMC) were processed through spatial matching, feature engineering, and geometric extraction to build robust predictive models. Various ML algorithms, including logistic regression, support vector machines, gradient boosting decision trees, and K-Means clustering, were used to assess and classify risks, with ensemble classifiers showing superior accuracy, especially when paired with Principal Component Analysis (PCA) for dimensionality reduction. Finally, a thorough data analysis highlighted spatial and operational factors influencing risks, identifying high-risk zones for focused monitoring. Overall, the study demonstrates the transformative potential of integrating GIS and ML in flowline risk management, proposing a data-driven approach that emphasizes the need for accurate data and refined models to improve safety in petroleum extraction.

\section*{Introduction}

As the global energy landscape evolves, the petroleum industry continues to grapple with the challenges of ensuring environmental safety and human well-being. In this work we lay the foundation for an extensive exploration of risk analysis of flowlines within the oil and gas sector. Our ultimate goal is to mitigate environmental impacts, improve safety and reduce human exposure associated with flowline failures through risk analysis using GIS and ML methods. However, a critical hurdle in this endeavor is the current lack of adequate data necessary to conduct such analysis. Therefore, this paper serves to provide a background on flowlines, review existing risk analysis literature, summarize how we approached cleaning and analyzing the data at hand, and outline possibilities for future work.

Flowlines, often overshadowed by more well studied pipelines, are crucial components in the petroleum production process. These underground conduits transport oil, natural gas, and water from wellheads to surface facilities and ultimately to Lease Automatic Custody Transfer (LACT) units. This research focuses on the `middle half' of the production process, where most U.S. flowlines are buried to prevent freezing and maintain structural integrity. Understanding the nuances of flowlines, including their material standards set by the American Petroleum Institute, construction, operational dynamics, and reasons for failure, is critical for a comprehensive risk assessment.

% Building on this understanding, the innovative approach of this research lies in the application of machine learning and GIS in assessing and mitigating risks associated with flowlines. 

Building on this understanding, the innovative approach of this research lies in combining the powerful tools of GIS and ML to build a risk model specific to flowlines. We implemented various techniques to determine which is best suited for this application. ML algorithms are adept at identifying patterns and predicting potential failures by analyzing vast datasets, which traditional methods might overlook. To take the first step in integrating multilinestring GeoDataFrames into ML models, we extracted key features from the flowlines to quantify the information they contain and incorporate it into the model. This feature extraction process allows the geometric and spatial properties of the flowlines to be represented in a way that the ML algorithms can interpret and use for improved predictive accuracy. The integrated method ultimately yields a more precise risk assessment model, enhancing the safety and reliability of the petroleum extraction process. However, a significant challenge in this effort is the lack of comprehensive data on flowline failures, which limits the depth and scope of the analysis and hinders the ability to fully predict the likelihood and impact of such failures. This introduction sets the stage for a detailed exposition of our research and data needs. We aim to provide a clear and comprehensive view of the complexities involved in flowline risk analysis and the transformative potential of ML and GIS in this field. Through this paper, we seek to engage our industry partners in a collaborative effort to gather the necessary data, paving the way for significant advancements in the safety and environmental stewardship of petroleum operations.

Before embarking on data exploration, we performed an extensive literature review to understand existing efforts in flowline and pipeline risk assessment. Several national and international organizations have developed models using diverse data types to predict risks associated with flowlines and pipelines. The primary goal of these models is to service or replace high-risk lines before they lead to spills, which can cause significant environmental damage and potentially result in loss of human life, especially in densely populated areas. This need has driven many organizations to seek out effective solutions to mitigate these risks. 

Risk assessment models have primarily been developed using two approaches: mathematical computation models and ML models. For instance, one mathematical model applied regression analysis to real-life data, calculating influence coefficients for each data input based on actual pipeline failures. It also assessed the potential damage of each pipeline's failure to prioritize high-risk pipelines (Vinogradov et al., 2018). On the other hand, various ML models have been employed, which are primarily data-driven rather than based on theoretical justifications. A study by Mazzella et al. (2019) compared three ML models, log-linear regression, eXtreme Gradient Boosting (xgBoost), and Artificial Neural Networks, to predict corrosion growth and found xgBoost to be the most accurate. Another study demonstrated the use of Euclidean-Support Vector Machines to predict pipeline failure using continuous sensor data (Lam Hong Lee et al., 2013). 

The literature shows that a combination of available data has been used to develop these models. The data's role is crucial in determining the reliability and accuracy of these risk models. The main categories of data used across the studies include pipeline specifications (diameter, age, coating, etc.), GIS data, soil data (for buried flowlines), human activity data (such as proximity to roads), inspection reports, repair/service history, historical incident records, operational data (flow rate, type of transported fluid, pressure, etc.), and continuous monitoring sensors data (for example, data from Long Range Ultrasonic Transducers) (Vinogradov et al., 2018, Mazzella et al., 2019, Zhang \& Liu, 2023, Khalilpasha \& Brown, 2023, Guan et al., 2019, Senouci et al., 2014, and Lam Hong Lee et al., 2013). Some studies employed simulated data as a proof of concept (Lam Hong Lee et al., 2013), while others used various combinations of the mentioned data categories to develop their risk models. The output of these models largely depends on the data used to train them. The literature primarily focuses on predicting corrosion, corrosion growth rate, risk ranking, remaining life, or type of failure. Flowline and pipeline failures typically stem from design issues, manufacturing flaws, installation errors, corrosion and erosion, structural threats (such as fatigue and static overload), natural hazards, and human error (Rachman et al., 2021). Predicting such outcomes is crucial for minimizing spill incidents, as it enables operators to proactively identify and address high-risk pipelines, staying one step ahead of potential failures. 
    
The effectiveness of a risk management model for flowlines and pipelines heavily depends on the accuracy, quantity, and diversity of the data used for training. Reliable data significantly improves the model's predictive capabilities. However, it's important to acknowledge that the model's capacity is limited to the data it's trained on. For example, a model trained on pipeline specifications, GIS data, and inspection reports might be highly accurate, but if it lacks training on human activity and operational data, it may miss failures linked to these factors. Therefore, careful selection of input data is a key factor in the model's success. When choosing data for the model, the relevance of features to failure determines the model's effectiveness, the cost of data acquisition should be sustainable, the reliability of the data must ensure an accurate model and the sufficiency of data must cover the entire area of interest.

\section*{Data Description and Processing}

The Colorado Energy and Carbon Management Commission (ECMC) maintains a range of datasets accessible on their website. Among these, the ``Spill Data" file was particularly pertinent to our study, encompassing individual cases for each spill recorded in Colorado since 2014. These cases are documented with an extensive array of variables, the specifics of which are thoroughly outlined on the ECMC website. To augment this spills dataset, ECMC provided two additional flowline datasets: the first, which we refer to as the descriptive data, includes GIS data pinpointing the exact locations of flowlines but lacks other details, starting with 259,979 flowlines. The second dataset, termed the operational data, details only the start and end points of flowlines but is enriched with significant operational data vital for constructing a risk model, initially containing 21,942 flowlines.

The initial step in our data processing was to amalgamate the descriptive and operational datasets into a unified flowline dataset. Due to the absence of unique identifiers, spatial characteristics were the basis for integration. The operational data were converted into a GeoDataBase (.gbd) format, with GPS coordinates adjusted to the ESPG:26913 coordinate reference system used in the descriptive data. We approximated each flowline by interpolating straight lines between provided endpoints and merged the two datasets by identifying intersections at both endpoints. Given the imprecision in the spatial data, a tolerance mechanism was implemented, starting at 0 and expanding up to 25 meters. If no match was found within this range, the line was excluded. This approach enabled us to successfully associate attributes from the operational data with the spatial geometry for 14,922 flowlines out of the initial 21,942, though the high density of flowlines and data inaccuracies potentially led to mismatches.

\textbf{Fig.~\ref{fig:1}} below illustrates this challenge. It depicts an approximated flowline (in red) from the operational data overlapping multiple precise flowlines (in blue) from the descriptive data. The lack of exact endpoint alignment complicates each match. To minimize errors, operator names were employed as an additional verification step; any mismatch in operator names, even within the tolerance range, led to the continuation of the search for a more accurate match. This, however, did not entirely eliminate inaccuracies, as often many of the flowlines originating from the same facility are maintained by the same operator, so overlaps in operator names across proximal flowlines are common.

\begin{figure}[h!]
\begin{subfigure}{0.5\textwidth}
  \centering
  \begin{frame}
  {\includegraphics[height = 1.5in, width=0.7\linewidth]{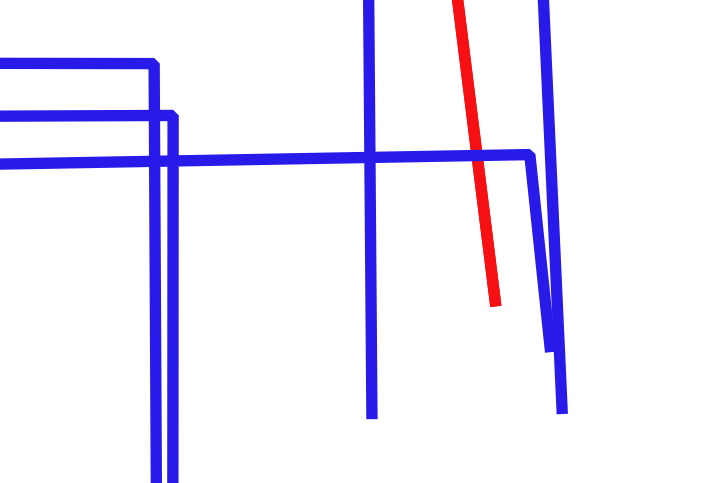}}
  \end{frame}
\end{subfigure}%
\begin{subfigure}{0.5\textwidth}
  \centering
  \begin{frame}
  {\includegraphics[height = 1.5in, width=0.7\linewidth]{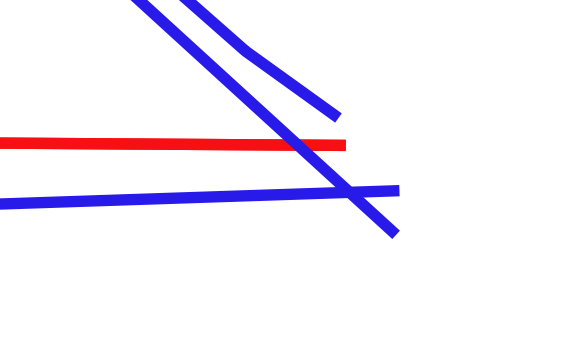}}
  \end{frame}
  \end{subfigure}
\caption{Patial Matching Challenges in Flowline Data Integration}
\label{fig:1}
\end{figure}
%
%
%\begin{center}\begin{frame}{\includegraphics[scale = 0.5506]{fig1aNEW.png}}\end{frame}\begin{frame}{\includegraphics[scale = 0.745]{fig1bNEW.png}}\end{frame}

\begin{figure}[h!]
\begin{center}
\begin{frame}
{\includegraphics[height = 1.5in, width=0.35\linewidth]{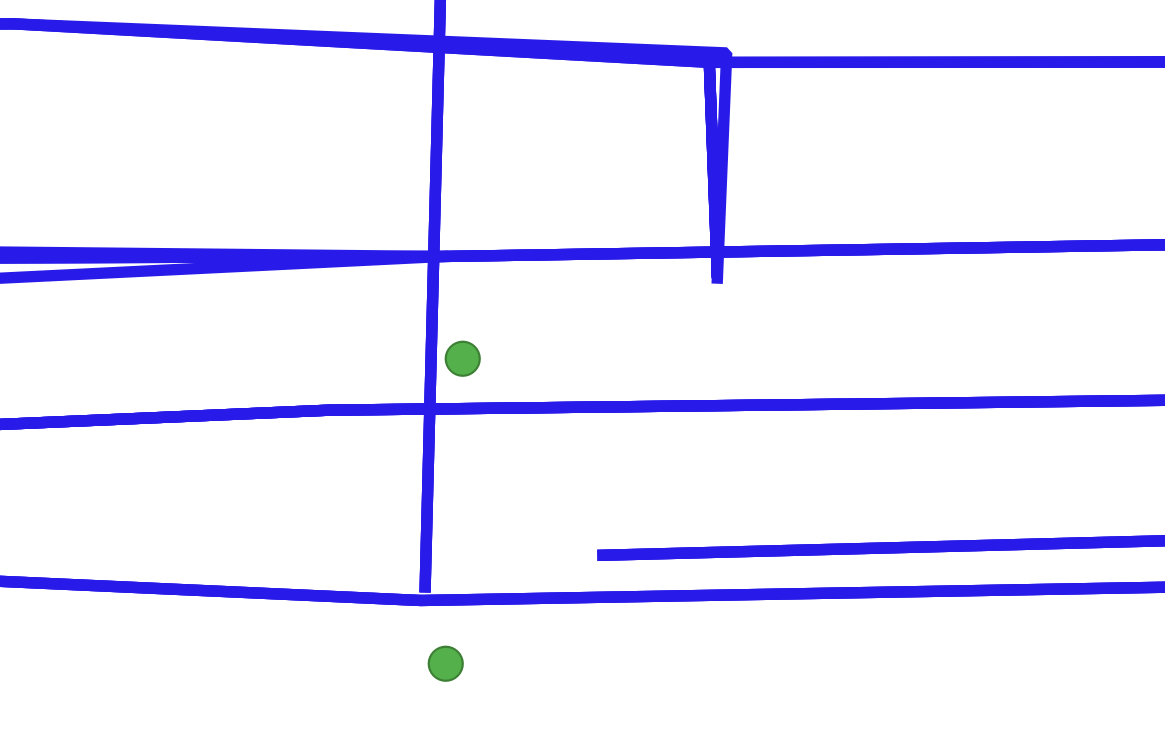}}
\end{frame}
\caption{Illustration of Potential Misalignment in Spill Matching}
\label{fig:2}
\end{center}
\end{figure}

%\small{Fig.~1: Spatial Matching Challenges in Flowline Data Integration}
%\end{center}
%
Subsequent to establishing a cohesive flowline dataset, we attempted to correlate each spill to its originating flowline. We initially identified 849 spills from the dataset, each represented as a point after conversion to the .gbd format. The spatial matching process was employed with an increasing tolerance from 0 to 25 meters to find the first intersecting flowline, using operator names for validation. However, the single-point comparison significantly heightened the risk of incorrect associations, resulting in only 84 spills being accurately matched to flowlines, as illustrated in \textbf{Fig.~\ref{fig:2}}.

\textbf{Fig.~\ref{fig:2}} depicts green points (spills) near blue lines (flowlines). The methodology assumed the nearest flowline with a matching operator name as the source, but this assumption was vulnerable to errors in spill location reporting. Often, a spill occurs underground, then moves laterally through the Earth, before reaching the surface where its location is recorded. In conjunction with measurement inaccuracy, this means it is plausible that the flowline from which a spill originated is not necessarily the closest flowline to the recorded location of the spill. Thus, a spill might be incorrectly matched to a nearby line rather than its true origin.

\section*{Feature Engineering}

Next, we concentrated on feature selection and engineering to refine our datasets for more effective analysis. One key attribute we developed was the age of each flowline, which we inferred from the construction date. Understanding the age of the flowlines is vital, as it might correlate with the likelihood of a spill occurring, under the hypothesis that older lines should be more prone to failures due to degradation over time. To differentiate flowlines based on their spill history, we introduced a binary risk column. This column labels each flowline as either high risk (1) if associated with a spill, or low risk (0) if no spills were recorded. This classification serves as a direct indicator for our analyses, helping to identify and prioritize flowlines that require more stringent monitoring and maintenance. To streamline our analysis, we meticulously selected variables that are essential for understanding flowline integrity and spill risks. Superfluous data that did not contribute to our risk assessment model were removed, allowing for a more focused and efficient analysis. A significant challenge in our data processing was addressing inconsistencies in the dataset, such as variations in spelling and formatting across various fields. Standardizing these elements was crucial for maintaining data integrity and ensuring accurate analysis. We tackled these discrepancies by mapping misspelled or inconsistently formatted variables to their correct versions. This mapping process was applied to multiple columns, ensuring that all entries under categories such as fluid type or operational status were uniformly formatted. This step was essential to prevent misclassification and errors during data processing, making the dataset cohesive and reliable for advanced statistical analyses and ML applications.

These preprocessing steps form a foundational part of our analysis by ensuring that the data is both accurate and relevant for modeling. The introduction of the line age and risk variables directly aligns with our study's goals to identify risk factors associated with flowlines. Pruning and standardizing the data simplifies the analytical process and enhances the robustness of our findings. As we move forward, these refined datasets are well-prepared for deployment in predictive modeling to effectively assess and predict spill risks. We hope that in detailing the challenges and steps of our data preparation, we can provide a framework for working with data that has similar limitations. 

To develop a ML model for predicting spill risks in flowlines, both datasets, flowlines with spills and without spills, were first merged. This combined dataset allowed us to analyze features consistently across all flowlines, enhancing the robustness of our predictive insights. Given that many of the attributes in our dataset are categorical without an inherent order (nominal data), using one-hot encoding was necessary. Attributes such as Operator Number, Flowline ID, Location ID, Status, Flowline Action, Location Type, Fluid Type, Material, and Root Cause Type were transformed using one-hot encoding. This technique prevents the introduction of arbitrary ordinality that numeric encoding might suggest. One-hot encoding converts each category value into a new binary column, ensuring that the model interprets these variables correctly without assuming a natural ordering.

Extracting meaningful features from the Geometry column, which is Geo-Spatial location data of the flowline, in our dataset was crucial since ML models cannot directly interpret geometric data. We decided to tabularize the data; extracting the most descriptive features to use in our model. This is the simplest and most intuitive approach to using GIS data as inputs for ML models. We derived three key features: Length, Number of Lines, and Bounding Box Area. The Length feature calculates the total length of each geometry. The Number of Lines feature quantifies the complexity of the geometry by counting the number of line segments it contains, differentiating between simple and complex structures. The Bounding Box Area is determined by calculating the area of the smallest rectangle that can completely enclose the geometry, providing insights into the spatial extent of each flowline. These features transform the geometric data into quantitative values that can be readily used in predictive modeling, improving our ability to assess and manage risks associated with the flowlines. The resulting attributes retained and engineered are seen in \textbf{Table~\ref{tab:attributes}}.

\begin{table}
\centering
\caption{Attributes for Flowline Analysis}
\label{tab:attributes}
\begin{tabular}{@{}ll@{}}
\toprule
\textbf{Attribute}                & \textbf{Description}                       \\ \midrule
Operator Number                   & Unique identifier for the operator         \\
Flowline ID                       & Unique identifier for the flowline         \\
Location ID                       & Identifier for the location of the flowline\\
Status                            & Operational status of the flowline         \\
Flowline Action                   & Actions taken or required on the flowline  \\
Location Type                     & Type of facility                            \\
Fluid Type                        & Type of fluid transported                  \\
Material                          & Construction material of the flowline      \\
Diameter (inches)                 & Diameter of the flowline in inches         \\
Length (feet)                     & Length of the flowline in feet             \\
Maximum Operating Pressure        & Max pressure flowline can withstand        \\
Shape Length                      & Geometrical length of the flowline         \\
Line Age (years)                  & Age of the flowline in years               \\
Length                            & Total length of each Geo-Spatial location  \\
Number of Lines                   & Number of line segments of Geo-Spatial location \\
Bounding Box Area                 & Boxed area of Geo-Spatial location          \\
Root Cause Type*                  & Underlying cause of the spill              \\ \bottomrule
\end{tabular}

\footnotesize{*Attribute included only for flowlines associated with spills.}
\end{table}

By transforming complex spatial data into these features, the dataset becomes more actionable for predictive modeling. These derived metrics provide insights into the physical characteristics and spatial distribution of the flowlines, which are instrumental in predicting and mitigating spill risks effectively. This thoughtful preparation and transformation of data ensure that the models built on this foundation are both accurate and reliable, crucial for making informed decisions in managing pipeline infrastructures. It is essential to recognize that our analysis is based on data that may contain inherent errors. This acknowledgment is critical when interpreting the results of our study, as it emphasizes the potential limitations of the data-driven insights we have derived. By admitting the possible imperfections in our data, we emphasize the need for cautious interpretation and validation of our predictive models, ensuring that decisions based on this research are made with a full understanding of these caveats. 
%\vspace{5mm}
\section*{Exploratory Data Analysis}

In this research, we first perform an exploratory data analysis (EDA) to uncover patterns, anomalies, and insights from spatial and frequency data related to risk assessment across Colorado. EDA serves as a fundamental phase in our study, allowing us to visually and statistically interpret the distribution of risk, which is crucial for subsequent predictive modeling and decision-making processes. By mapping out risk levels and their distribution, we aim to identify high-risk areas that require immediate attention and better understand the underlying factors contributing to these risks. This approach not only improves our understanding of the spatial dynamics but also facilitates a targeted response strategy based on empirical evidence.

\begin{figure}[h!]
\begin{center}
\begin{frame}
{\includegraphics[height = 4in, width=0.8\linewidth]{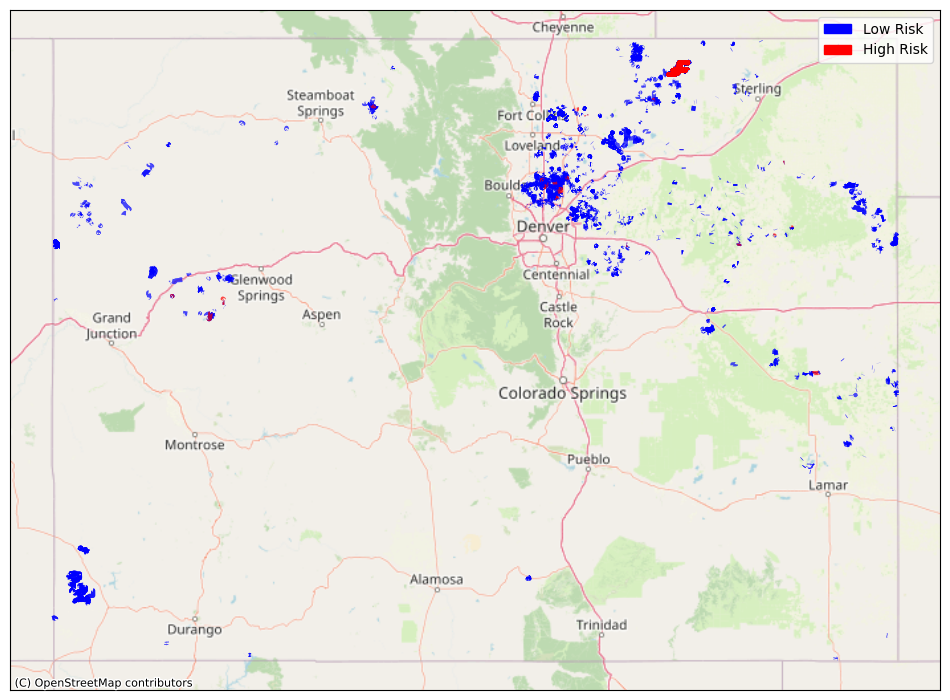}}
\end{frame}
\caption{Spatial Distribution of Flowline Risk in Colorado}
\label{fig:3}
\end{center}
\end{figure}

\textbf{Fig.~\ref{fig:3}} provides a visual representation of risk distribution across Colorado. Blue markers indicate locations classified as `Low Risk', while red markers denote `High Risk' areas. The concentration of blue markers across the state suggests that most regions are currently at low risk. However, there are notable clusters of high-risk areas, particularly around major urban centers such as Denver. This spatial distribution is essential for regional planning and resource allocation, highlighting areas that might require more intensive monitoring and preventive measures. 

\section*{Risk Assessment by Operational Parameters}
\begin{figure}[h!]
\centering
\begin{subfigure}{0.49\textwidth}
    \includegraphics[width=\linewidth]{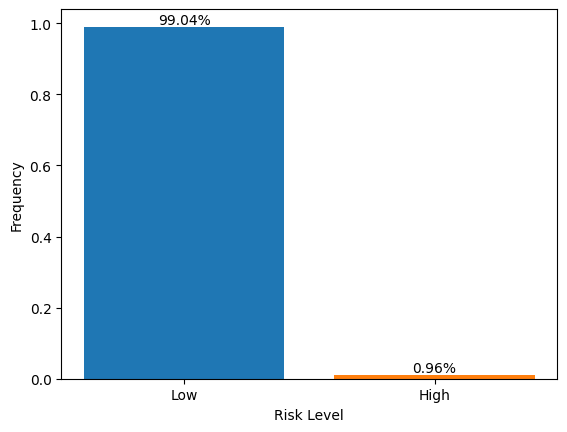}
    \caption{Frequency Distribution of Risk Levels}
    \label{fig:high-low}
\end{subfigure}
\hfill
\begin{subfigure}{0.49\textwidth}
    \includegraphics[width=\linewidth]{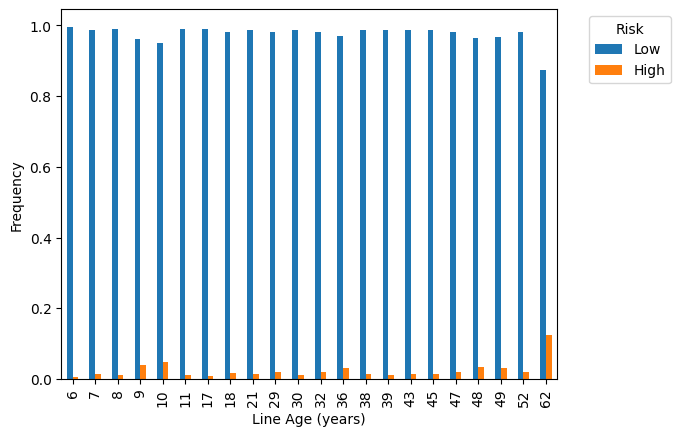}
    \caption{Risk vs. Line Age}
    \label{fig:risk-age}
\end{subfigure}
\vspace{5mm} % adds vertical space between the rows

\begin{subfigure}{0.49\textwidth}
    \includegraphics[width=\linewidth]{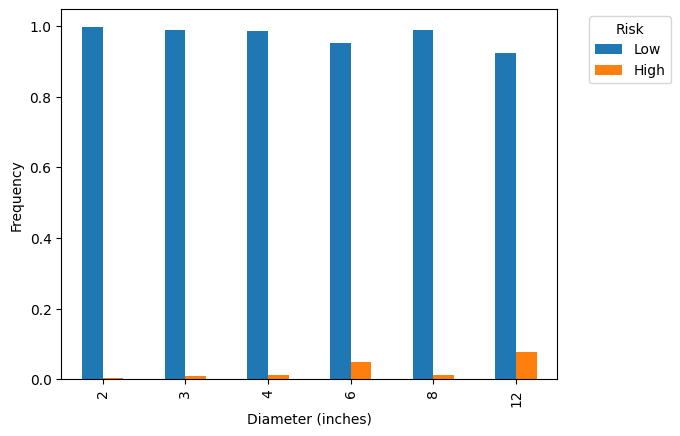}
    \caption{Risk vs. Diameter}
    \label{fig:risk-diameter}
\end{subfigure}
\hfill
\begin{subfigure}{0.49\textwidth}
    \includegraphics[width=\linewidth]{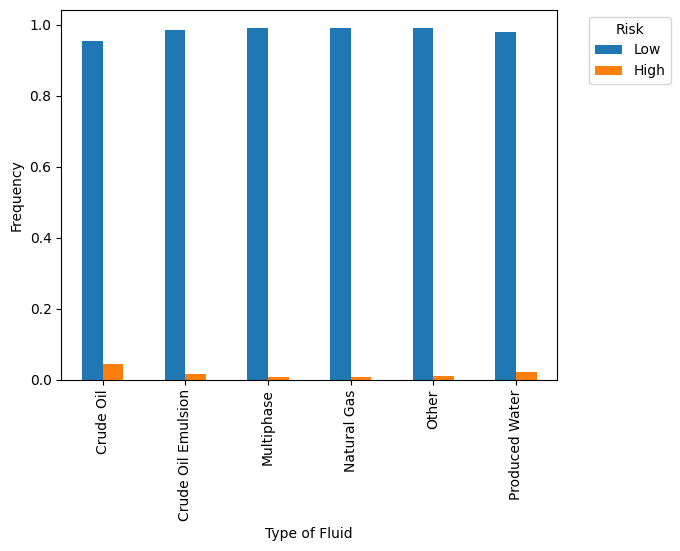}
    \caption{Risk vs. Type of Fluid}
    \label{fig:risk-fluid}
\end{subfigure}
\vspace{5mm} % adds vertical space between the rows

\begin{subfigure}{0.49\textwidth}
    \includegraphics[width=\linewidth]{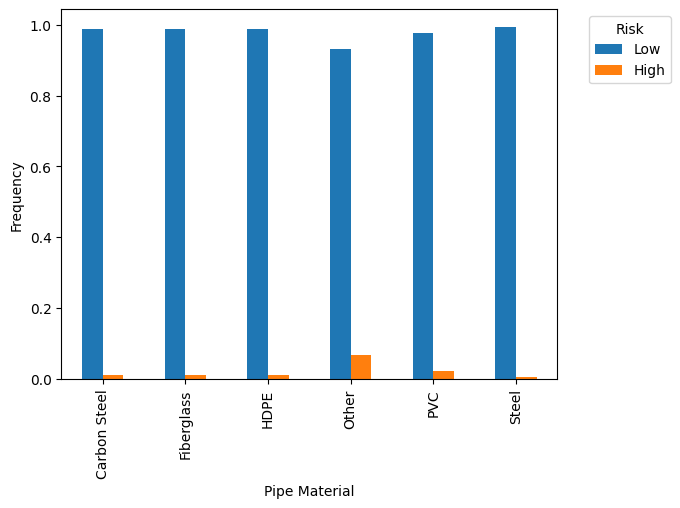}
    \caption{Risk vs. Pipe Material}
    \label{fig:risk-material}
\end{subfigure}
\hfill
\begin{subfigure}{0.49\textwidth}
    \includegraphics[width=\linewidth]{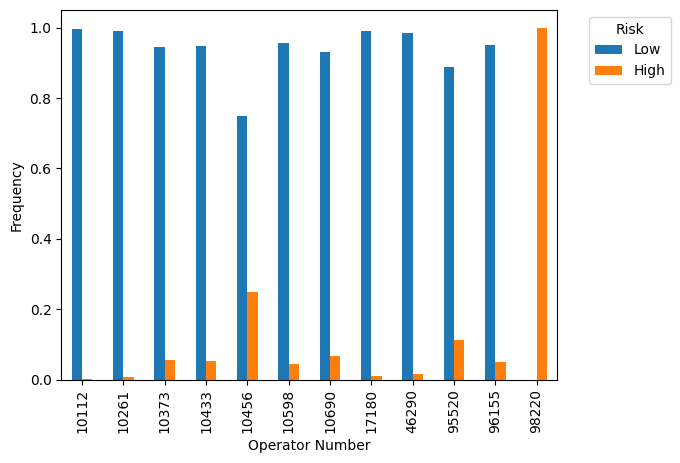}
    \caption{Risk vs. Operator Number}
    \label{fig:risk-operator}
\end{subfigure}

\caption{Various aspects of risk analysis}
\label{fig:risk-analysis}
\end{figure}

The following graphs illustrate the relationship between various operational parameters and the associated risk levels of pipeline systems in Colorado. Each graph explores a different aspect: line age, pipe diameter, type of fluid, pipe material, and operator number. \textbf{Fig.~\ref{fig:high-low}} is a bar chart that quantifies the proportion of locations at low and high risk levels. Approximately 99.04\% of the observed locations are categorized under `Low Risk', while only 0.96\% fall under `High Risk'. This visualization simplifies the understanding of risk prevalence across the state, underlining the rarity but potential severity of high-risk zones. This stark contrast in frequency underscores the importance of focusing on and managing these high-risk areas to mitigate potential adverse outcomes effectively. \textbf{Fig.~\ref{fig:risk-age}} shows the frequency of low and high risk across different line ages in years. It is clear that the risk remains predominantly low across all ages, with minor instances of high risk noted at greater line ages. This suggests that older lines might incrementally contribute to risk, albeit minimally. The risk distribution across various pipe diameters is shown in \textbf{Fig.~\ref{fig:risk-diameter}}. Similar to line age, the vast majority of lines, regardless of diameter, are categorized as low risk. However, there is a small increase in high risk observed in larger diameters (8 in. and 12 in.). This may suggest that larger pipes, which are likely subject to higher pressure or greater fluid velocity (rather than volume alone), could pose a higher risk factor.
\textbf{Fig.~\ref{fig:risk-fluid}} categorizes risk level by the type of fluid transported. The fluids include crude oil, multiphase, natural gas, other, and produced water. Across all fluid types, the risk remains overwhelmingly low with a negligible proportion of high-risk instances for all fluid types apart from crude oil. It appears that flowlines transporting crude oil pose a slightly higher risk. In \textbf{Fig.~\ref{fig:risk-material}}, different pipe materials such as carbon steel, fiberglass, HDPE, other, PVC, and steel are examined for risk levels. Across all materials, there's a consistent predominance of low risk. However, certain materials like HDPE and steel show a slightly higher occurrence of high risk, potentially indicating material-specific vulnerabilities, though other factors such as fluid type, age, or pressure may also contribute. In \textbf{Fig.~\ref{fig:risk-operator}}, analyzing risk by operator number reveals that most operators manage their lines with low risk. A notable increase in high-risk frequency is observed with one particular operator (operator number 98220), highlighting the possibility of operator-specific practices influencing risk levels more significantly than other factors.

\section*{\textbf{Methodology}}

In our study, we performed risk analysis using both supervised and unsupervised ML techniques, with an optional application of Principal Component Analysis (PCA) for data reduction. This methodology was structured to address the complexities of predicting spill risks in flowlines by leveraging various ML models to explore different aspects of the dataset (Hastie et al., 2009). To prepare our dataset for ML applications, we first isolated the predictors (`inputs') and the outcome variable (`outputs'). The `inputs' included all features of interest from our dataset except for `risk', which do not directly serve as inputs for modeling. The `outputs', specifically the `risk' column, were designated as our target variable, indicating the presence or absence of a spill. To address challenges associated with high-dimensional data, we performed PCA. This technique is particularly beneficial for reducing the dimensionality of a dataset while retaining the most informative features that capture the maximum variance. By doing so, PCA helps to simplify the model, reduce overfitting, and potentially improve predictive accuracy by focusing solely on the principal components.

The division of the dataset into training and testing sets was done through a stratified split, ensuring that both subsets represent the overall distribution of the data accurately. We allocated 70\% of the data for training and 30\% for testing. This split was critical for training our models on a sample of data while reserving a separate portion for unbiased evaluation of model performance.

\vspace{5mm}

\section*{\textbf{Supervised and Unsupervised Learning Techniques}}

In our analysis of spill risks in flowlines, we employed a diverse array of ML models, categorizing them into supervised and unsupervised methods. Within the supervised category, we further divided the techniques into single classifiers and ensemble classifiers, each with specific strengths and approaches to handling the prediction tasks. For the single classifiers, we chose three methods. Logistic regression (LR) is a fundamental statistical approach that models the probability of a binary outcome based on one or more predictor variables (Hastie et al., 2009). It is particularly useful for understanding the impact of several independent variables on a binary response, making it ideal for binary classification tasks such as predicting whether a spill will occur or not. K-Nearest Neighbors (K-NN) is a non-parametric method used for classification by comparing the feature similarity. A data point is classified by a majority vote of its neighbors, with the data point being assigned to the class most common among its $k$ nearest neighbors (Hastie et al., 2009). K-NN is highly intuitive and straightforward but can become computationally expensive as the size of the data grows. Support Vector Machine (SVM) is a powerful classification technique that finds the hyperplane that best divides a dataset into classes. It is effective in high-dimensional spaces and robust against overfitting, especially in complex domains where the margin between different classes is clear (Hastie et al., 2009). 

For the ensemble classifiers, we also chose three methods. Gradient Boosting Decision Trees (GBDT) is an ensemble technique that builds the model in a stage-wise fashion. It constructs new models that predict the residuals or errors of prior models and then combines them into a final model (Hastie et al., 2009). It’s often used for its high performance and predictive power, particularly in competitions and real-world applications. Adaptive Boosting (AdaBoost) is one of the first boosting algorithms to be adapted to solving practices. It works by combining multiple weak classifiers to create a strong classifier. During the training phase, AdaBoost assigns weights to each instance, which are adjusted as each successive model is built, focusing more on difficult to classify instances (Hastie et al., 2009). Random Forests (RF) construct a multitude of decision trees at training time and output the class that is the mode of the classes (classification) of the individual trees. It is highly versatile and capable of handling both regression and classification tasks, offering a good balance between accuracy and computational efficiency (Breiman, 2001).

For unsupervised learning, we chose K-Means Clustering. K-Means is a popular clustering algorithm that groups data points into a predefined number of clusters ($k$) based on their feature similarities. It iteratively assigns each data point to one of $k$ groups based on the nearest mean. The algorithm minimizes the variance within each cluster, making it effective for identifying distinct groups or patterns in the data (Hastie et al., 2009). It is particularly useful for market segmentation, anomaly detection, and other applications where the data labels are unknown. 

Each of these methods brings different strengths and weaknesses to the table. By integrating multiple approaches, from simple models like logistic regression to more complex ensemble methods, and the exploratory power of K-Means clustering, we aim to capture a broad spectrum of insights from the dataset, providing a robust analytical framework for assessing and predicting spill risks effectively.

\section*{\textbf{Results}}

In ML, particularly in classification tasks, several key performance metrics are crucial for evaluating the effectiveness of models. These metrics include accuracy, precision, recall, and the F1 score, each offering unique insights into a model's performance.

Accuracy measures the proportion of total correct predictions (both true positives and true negatives) among all cases evaluated. It provides a straightforward indicator of a model's overall effectiveness across all classes. However, its reliability can diminish in the presence of unbalanced datasets where one class significantly outnumbers another. Precision assesses the accuracy of positive predictions, that is, the proportion of true positive results among all positive predictions made by the model. High precision is vital in scenarios where the consequences of false positives are significant, ensuring that when a model predicts a risk, such a prediction is likely to be correct. Recall, or sensitivity, measures the model's ability to identify all actual positives from the data. It represents the proportion of true positives identified relative to the total actual positives. This metric is critical in situations where missing a positive instance carries serious consequences, such as failing to identify a potential hazard in risk management systems. The F1 Score harmonizes precision and recall by calculating their harmonic mean. This metric is particularly useful when seeking a balance between precision and recall, especially in environments with uneven class distributions. It ensures that a model does not overly favor one metric over the other, providing a more balanced view of model performance. 

In the context of this study on flowline risk in the oil and gas industry, these metrics collectively inform the robustness of various ML models used to predict spill risks. High scores in these metrics indicate a model's capability to accurately predict and manage potential risks, crucial for ensuring safety and operational efficiency. The results from the ML models used in our study to predict spill risks in flowlines reveal significant insights into the effectiveness of each classifier under different circumstances, such as before and after applying PCA.

\section*{\textbf{Without PCA Reduction}}

\textbf{Table~\ref{table:2}} shows the results of our supervised ML model without PCA reduction. LR and SVM both display exceptionally high performance across accuracy, precision, recall and mean F1 score. This suggests that these models are highly effective in predicting the binary outcome of spill occurrences when the full feature set is considered. K-NN, while also high in accuracy, shows significantly lower precision, recall, and F1 scores, indicating it may be less effective or consistent in handling this particular data distribution or feature set. Among the ensemble classifiers, GBDT and AdaBoost show robust performance but lower recall and F1 scores compared to single models like LR and SVM. The RF classifier shows a drop in recall and F1 score, pointing to challenges in managing the balance between detecting true positives and avoiding false negatives.

\begin{table}[ht]
\centering
\caption{Performance Metrics of Classifiers}
\label{table:2}
\begin{tabular}{@{}lcccc@{}} % Specifies left alignment for text and center for numbers, adjust as necessary
\toprule
Classifier & Accuracy & Precision & Recall & Mean F1 Score \\
\midrule
\multicolumn{5}{c}{Single Classifiers} \\
\midrule
LR & 0.990 & 0.99 & 0.99 & 0.99 \\
K-NN & 0.992 & 0.80 & 0.33 & 0.47 \\
SVM & 0.990 & 0.99 & 0.99 & 0.99 \\
\midrule
\multicolumn{5}{c}{Ensemble Classifiers} \\
\midrule
GBDT & 0.995 & 0.89 & 0.67 & 0.76 \\
AdaBoost & 0.995 & 0.89 & 0.67 & 0.76 \\
RF & 0.993 & 0.99 & 0.33 & 0.50 \\
\bottomrule
\end{tabular}
\end{table}

\section*{\textbf{With PCA Reduction}}

\textbf{Table~\ref{table:3}} shows the results of our supervised maching learning model with PCA reduction. The performance of LR drastically changes, showing a steep decline in all metrics, which indicates that PCA may have removed features critical to the model’s prediction capabilities for LR. K-NN and SVM maintain high accuracy, but K-NN shows a dramatic drop in recall and F1 scores, suggesting it struggles with the reduced feature set's ability to capture the necessary proximity or similarity among data points for effective classification. Ensemble methods like GBDT and AdaBoost remain robust even after PCA, maintaining high scores across all metrics. This suggests that these methods are well-suited to handle the dimensional reduction without losing key information needed for classification. RF experiences a reduction in all performance metrics, indicating possible overfitting with the full feature set or loss of critical information for classification in the reduced feature set.

\begin{table}[H]
\centering
\caption{Performance Metrics of Classifiers with PCA Reduction}
\label{table:3}
\begin{tabular}{@{}lcccc@{}} % Specifies left alignment for text and center for numbers, adjust as necessary
\toprule
Classifier & Accuracy & Precision & Recall & Mean F1 Score \\
\midrule
\multicolumn{5}{c}{Single Classifiers} \\
\midrule
LR & 0.733 & 0.02 & 0.47 & 0.04 \\
K-NN & 0.989 & 0.99 & 0.13 & 0.24 \\
SVM & 0.987 & 0.99 & 0.99 & 0.99 \\
\midrule
\multicolumn{5}{c}{Ensemble Classifiers} \\
\midrule
GBDT & 0.987 & 0.99 & 0.99 & 0.99 \\
AdaBoost & 0.987 & 0.99 & 0.99 & 0.99 \\
RF & 0.986 & 0.40 & 0.27 & 0.32 \\
\bottomrule
\end{tabular}
\end{table}

\section*{\textbf{Unsupervised Model}}

\begin{figure}[H]
    \centering
    \includegraphics[width=0.6\linewidth]{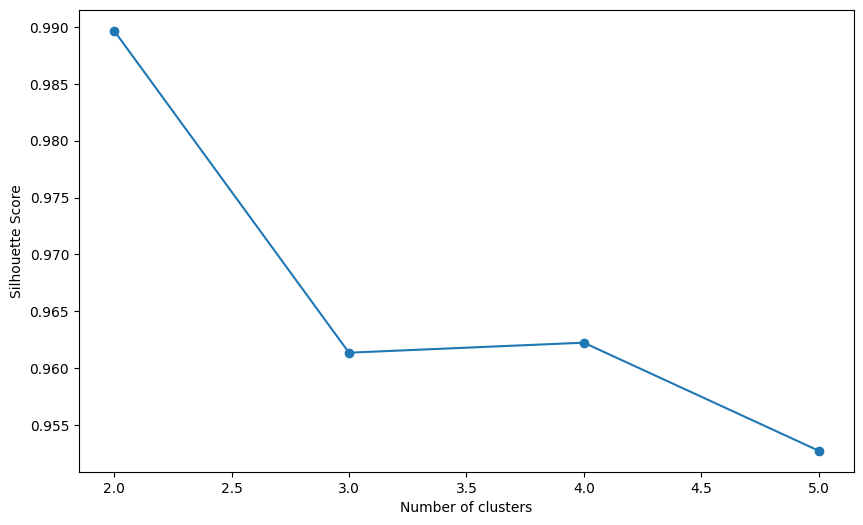}
    \caption{Silhouette Score by Cluster Number}
    \label{fig:5}
\end{figure}

For our unsupervised model, we used K-means Clustering. In \textbf{Fig.~\ref{fig:5}}, we see the results of the model's Silhouette Score; a measure of how similar an object is to its own cluster compared to other clusters. The graph provided shows the Silhouette Score for different numbers of clusters ranging from 2 to 5. The highest Silhouette Score is observed when the number of clusters is 2, which is close to 0.990. This high score suggests that the two-cluster solution provides a good fit, meaning that the data points within each cluster are quite similar to each other and relatively dissimilar to points in other clusters. As the number of clusters increases from 2 to 5, there is a noticeable decline in the Silhouette Score, indicating that the additional clusters may be forcing dissimilar points together or splitting naturally similar points into different groups. The substantial drop when moving from 2 to 3 clusters and a continued decrease up to 5 clusters suggests that increasing the number of clusters beyond 2 leads to less cohesive and more dispersed clusters.

\begin{figure}[H]
    \centering
    \includegraphics[width=0.6\linewidth]{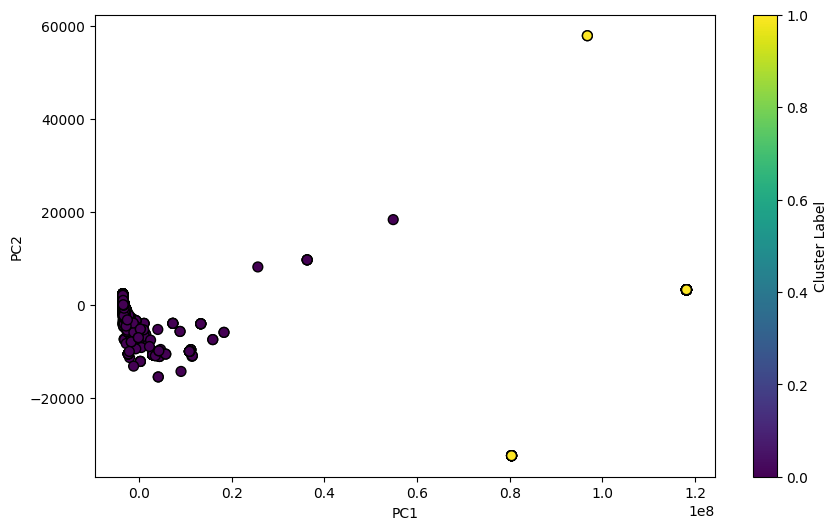}
    \caption{Predicted: PCA of Dataset with K-means Clustering}
    \label{fig:6}
\end{figure}

\begin{figure}[H]
    \centering
    \includegraphics[width=0.6\linewidth]{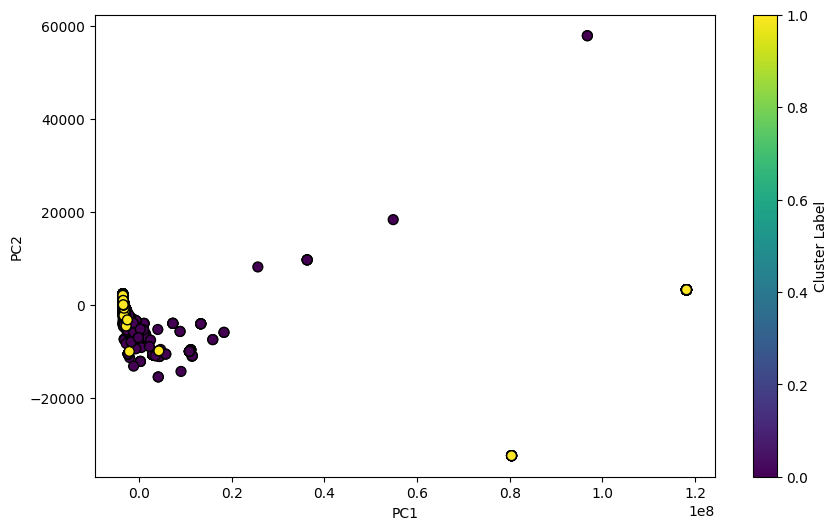}
    \caption{Actual: PCA of Dataset with K-means Clustering}
    \label{fig:7}
\end{figure}

In \textbf{Fig.~\ref{fig:6}} and \textbf{Fig.~\ref{fig:7}}, the PCA visualizations for actual and predicted clustering help illustrate how the dataset characteristics are captured by PCA and how effectively K-Means clustering can group these reduced features. Both actual and predicted PCA plots show the data points spread primarily along the first principal component (PC1), indicating that PC1 captures the most significant variance within the dataset. The spread along PC2 is considerably less, which points to its lower significance in variance compared to PC1. In the PCA plots, we observe that most data points are concentrated in a dense cluster with some outliers. The actual PCA plot reveals a denser clustering around the center, while the predicted PCA plot shows a slightly more dispersed distribution, especially at higher values of PC1.

The color gradient (from purple to yellow) represents different clusters. In the actual PCA, there is a clear distinction between two major clusters, with the yellow cluster having fewer points, likely representing an outlier or a specific subgroup within the dataset. The predicted PCA also shows a similar pattern, though the separation between clusters appears less distinct, suggesting some degree of error or variation in how the K-Means algorithm has assigned cluster labels based on the PCA-reduced features.

\section*{\textbf{Conclusions}}

The key findings from our research demonstrate the significant potential of integrating ML and Geographic Information Systems (GIS) in the risk analysis of flowlines within the oil and gas industry. Despite the challenges of our dataset, and the tabularization of GIS data features, which loses some spatial information, we were able to show the potential that this approach has for extremely accurate risk modeling. One of the novel contributions of this work is the exploration of multilinestring GeoDataFrames in ML models. While GeoDataFrames are more conventionally used as raster data in ML models, such as satellite data, the use of multilinestring data has not been extensively explored. In this study, we developed a preliminary method for integrating multilinestring data by extracting quantifiable features, allowing the ML models to interpret and use this spatial information effectively. This opens up new possibilities for incorporating more complex geospatial data into ML models.

The EDA revealed crucial insights into the spatial distribution of risk, identifying low-risk zones and pinpointing high-risk areas requiring urgent attention. This analysis was instrumental in setting the stage for deeper ML analysis by highlighting the regions and parameters most associated with risk. The integration of GIS provided a spatial analysis platform that improved the interpretation of data by relating geographical variables with flowline integrity. This integration is pivotal for understanding environmental factors affecting flowlines and allows for targeted interventions tailored to specific geographic conditions.

Our results from supervised learning models, both with and without PCA reduction, indicate a varying degree of effectiveness across different classifiers. LR and SVM displayed exceptional performance across all metrics without PCA, signifying their robustness in handling full feature sets. However, the introduction of PCA, which simplifies the data by reducing dimensions, adversely affected some models, particularly LR and K-Nearest Neighbors, highlighting the complexity of feature selection in model accuracy. The unsupervised model using K-Means clustering highlighted the optimal clustering of data into two distinct groups, which was evident from the Silhouette Scores. This finding was critical as it demonstrated the natural grouping within the data, which could potentially categorize flowline systems by risk level effectively.

Despite the successes, our study faces limitations, primarily related to data availability and quality. The need for comprehensive, accurate, and accessible data remains a significant challenge that restricts the depth and scope of potential analyses. Future efforts will focus on expanding data collection and refining ML models to improve predictive accuracy. Additionally, ongoing collaborations with industry partners will be crucial to gather expansive datasets that reflect a wider range of operational conditions and environmental factors.

In future we plan to incorporate digital elevation models (DEM) along with more advanced segmentation methods to further refine our multilinestring analysis. Specifically, we will segment the multilinestrings into single line strings where we have the most accurate location of spills, allowing for more precise predictions of where spills are likely to occur. Additionally, we are investigating methods to retain geolocation within the ML models so that we can spatially map the predictions, which would enable more detailed and actionable risk assessments.

The implications of this study are profound for the oil and gas industry. By improving the predictive accuracy of risk assessments, our research supports proactive risk management strategies that can prevent environmental disasters and protect human lives. Furthermore, the methodologies developed here contribute to the scientific community by providing a framework for integrating ML with traditional risk analysis techniques in other sectors. In conclusion, this study represents a significant step forward in the use of advanced technologies to address critical safety and environmental challenges in the oil and gas industry. The combination of ML, GIS, and risk analysis methodologies offers a new horizon for predictive analytics in industrial applications. As we continue to refine these techniques and expand our datasets, the potential to transform risk management practices and improve safety standards globally becomes increasingly attainable.

Finally, the code required to reproduce all the results presented in this manuscript is publicly available at \url{https://github.com/ichittumuri/SPE\_Journal\_Risk\_Analysis\_of\_Flowlines}.

\section*{Nomenclature}

\begin{itemize} \item \textbf{AdaBoost} = Adaptive Boosting – an ensemble learning method for improved accuracy \item \textbf{DEM} = Digital Elevation Model – a representation of terrain surface \item \textbf{ECMC} = Colorado Energy and Carbon Management Commission – the data source in this study \item \textbf{ESPG:26913} = A specific Coordinate Reference System (CRS) for mapping \item \textbf{GBDT} = Gradient Boosting Decision Trees – an ensemble classifier used in risk analysis \item \textbf{GIS} = Geographic Information Systems – used to analyze and visualize spatial data \item \textbf{K-Means} = K-Means Clustering – an unsupervised ML algorithm \item \textbf{K-NN} = K-Nearest Neighbors – a non-parametric classification method \item \textbf{LR} = Logistic Regression – a binary classification model used in this study \item \textbf{ML} = Machine Learning – used to predict risk in flowlines \item \textbf{PCA} = Principal Component Analysis – used to reduce data dimensionality \item \textbf{RF} = Random Forest – an ensemble method used in classification tasks \item \textbf{SVM} = Support Vector Machine – a classification algorithm used for flowline risk analysis \end{itemize}

\section*{\textbf{Acknowledgments}}
This work was supported by the The Mark Martinez and Joey Irwin Memorial Public Projects Fund: Design and Statistical
Analysis of a Flowline Risk Assessment Model. The authors would like to acknowledge ECMC for providing the data essential for this study. Special thanks to Matthew Bauer, an  Affiliate Faculty from the Colorado School of Mines, for his valuable GIS contributions and support during the research process.

\bibliographystyle{plain}

\begin{thebibliography}{9}

\bibitem{senouci2014}
Ahmed Senouci, Mohamed Elabbasy, Emad Elwakil, Bassem Abdrabou, and Tarek Zayed, ``A model for predicting failure of oil pipelines," \textit{Structure and Infrastructure Engineering}, vol. 10, no. 3, pp. 375-387, 2014. DOI: 10.1080/15732479.2012.756918.

\bibitem{rachman2021}
Andika Rachman, Tieling Zhang, and R.M. Chandima Ratnayake, ``Applications of machine learning in pipeline integrity management: A state-of-the-art review," \textit{International Journal of Pressure Vessels and Piping}, vol. 193, 2021. https://doi.org/10.1016/j.ijpvp.2021.104471.

\bibitem{cogcc2019}
Colorado Oil \& Gas Conservation Commission, ``Annual Flowline Spill Report - 2019," 22Denver, Colorado, USA, January 1, 2019 – December 30, 2019.

\bibitem{khalilpasha2023}
Hossein Khalilpasha and Justin Brown, ``Minimising Pipeline Leaks and Maximising Operational Life by Application of Machine Learning at Cooper Basin," presented at the AMPP Annual Conference + Expo, Denver, Colorado, USA, March 2023.

\bibitem{mazzella2019}
Joseph Mazzella, Thomas Hayden, Len Krissa, and Haralampos Tsaprailis, ``Estimating Corrosion Growth Rate for Underground Pipeline: A Machine Learning Based Approach," presented at the CORROSION 2019, Nashville, Tennessee, USA, March 2019.

\bibitem{lee2013}
Lam Hong Lee, Rajprasad Rajkumar, Lai Hung Lo, Chin Heng Wan, and Dino Isa, ``Oil and gas pipeline failure prediction system using long range ultrasonic transducers and Euclidean-Support Vector Machines classification approach," \textit{Expert Systems with Applications}, vol. 40, no. 6, pp. 1925-1934, 2013. https://doi.org/10.1016/j.eswa.2012.10.006.

\bibitem{vinogradov2018}
P. V. Vinogradov, K. V. Litvinenko, R. I. Valiakhmetov, and A. N. Bakhtegareeva, ``Development of a model for ranking field pipelines based on risk assessment in exploitation," \textit{OIJ}, vol. 2018, pp. 84–86, 2018. doi: 10.24887/0028-2448-2018-8-84-86.

\bibitem{guan2019}
Shan Guan, Francois Ayello, Narasi Sridhar, Xiaoming Han, Qingshan Feng, and Yonghe Yang, ``Application of Probabilistic Model in Pipeline Direct Assessment," presented at the CORROSION 2019, Nashville, Tennessee, USA, March 2019.

\bibitem{Fleckenstein2018}
William Fleckenstein, ``Flowline Risk Review – Final Report," presented to Mark Schlagenhauf and Stuart Ellsworth, Colorado School of Mines, Golden, Colorado, USA, October 2018.

\bibitem{zhang2023}
Zhuoran Zhang and Guanlan Liu, ``A Prediction of Corrosion-Related Leakage on Distribution Pipelines via Machine Learning Method," presented at the AMPP Annual Conference + Expo, Denver, Colorado, USA, March 2023.

\bibitem{breiman2001}
L. Breiman, ``Random Forests," \textit{Machine Learning}, vol. 45, no. 1, pp. 5–32, 2001. doi: 10.1023/A:1010933404324.

\bibitem{hastie2009}
T. Hastie, R. Tibshirani, and J. Friedman, \textit{The Elements of Statistical Learning: Data Mining, Inference, and Prediction}, 2nd ed. New York, NY, USA: Springer, 2009. doi: 10.1007/978-0-387-84858-7.

\end{thebibliography}

\vspace{5mm}

%\section*{\textbf{Figures and Tables (See SPE Style Guide, Sections 2.10 and 2.11, Appendices H and I)}} 

%For final submission after acceptance, tables and all figure captions should be moved from within the text to the end of the paper.

\end{document}